\documentclass[11pt,a4paper,xcolor=table]{article}
\usepackage[table,xcdraw]{xcolor}
\usepackage[hyperref]{emnlp2020}
\usepackage{times}
\usepackage{latexsym}
\usepackage{amsfonts}

\usepackage[section]{placeins}

\usepackage{microtype}
\usepackage{graphicx} 
\usepackage{float} 
\usepackage{subfigure} 
\usepackage{multirow}
\usepackage{booktabs}

\usepackage[linesnumbered, ruled]{algorithm2e}
\usepackage[T1]{fontenc}
\usepackage{dsfont}
\usepackage{colortbl}
\usepackage{xcolor}
\usepackage{enumitem}

\usepackage{mdframed}
\usepackage{booktabs}
\usepackage{graphicx}

\usepackage[normalem]{ulem}

\usepackage{xcolor}
\usepackage{soul}
\colorlet{soulred}{red!30}
\sethlcolor{soulred}%
\usepackage{amsmath,amsthm,amsfonts,amssymb,bm}

\usepackage{framed}
\usepackage{varioref}
\usepackage{float}
\usepackage{multirow}
\usepackage{dashrule}
\usepackage{xurl}
\usepackage{pifont}

\aclfinalcopy 
\def\aclpaperid{2473} 

\title{Improving Event Duration Prediction \\ via Time-aware Pre-training}

\author{Zonglin Yang\qquad Xinya Du\qquad Alexander Rush\qquad Claire Cardie
\\Department of Computer Science
\\Cornell University
\\ {\tt \{zy223, arush\}@cornell.edu}
\\ {\tt \{xdu, cardie\}@cs.cornell.edu}}

\date{}

\begin{document}

\maketitle
\begin{abstract}

End-to-end models in NLP rarely encode external world knowledge about length of time.
We introduce two effective models for duration prediction, which incorporate external knowledge by reading temporal-related news sentences (time-aware pre-training).
Specifically,
one model predicts the \textit{range/unit} where the duration value falls in (\textsc{R-pred}); and the other predicts the \textit{exact} duration value (\textsc{E-pred}).
Our best model -- \textsc{E-pred}, substantially outperforms previous work, and captures duration information more accurately than \textsc{R-pred}.
We also demonstrate our models are capable of duration prediction in the unsupervised setting, outperforming the baselines.

\end{abstract}

\section{Introduction}
Understanding duration of event expressed in text is a crucial task in NLP \cite{pustejovsky2009semeval, zhou-etal-2019-going}. It facilitates downstream tasks such as story timeline construction~\cite{ning-etal-2018-multi, leeuwenberg2019survey} and temporal question answering~\cite{llorens2015semeval}. 
It is challenging to make accurate prediction mainly due to two reasons:
(1) duration is not only associated with event word but also the context. For example, ``watch a movie'' takes around 2 hours, while ``watch a bird fly'' only takes about 10 seconds; 
(2) the \textit{compositional nature} of events makes it difficult to train a learning-based system only based on hand annotated data (since it's hard to cover all the possible events). Thus, external knowledge and commonsense are needed to make further progress on the task.

However, most current approaches~\cite{pan-etal-2011-annotating, gusev2011using, vempala-etal-2018-determining} focus on developing features and cannot utilize external textual knowledge. 
The only exception is the web count based method proposed by~\newcite{gusev2011using}, which queries search engine with event word (e.g., ``watch'') and temporal units,
and make predictions based on hitting times. 
However, this method achieves better performance when query only with the event word
in the sentence, which means it does not enable contextualized understanding.

To benefit from the generalizability of learning-based methods and utilizing external temporal knowledge, we introduce a framework, which includes
(1) a procedure for collecting duration-related news sentences, and automatic labeling the duration unit in it (Section~\ref{Collecting data});
\footnote{We'll release these weakly supervised duration-relevant sentences in \url{https://github.com/ZonglinY/Improving-Event-Duration-Prediction-via-Time-aware-Pre-training.git}}
(2) two effective end-to-end models that leverage external temporal knowledge via pre-training (Section~\ref{method and pretraining}).
Specifically, our first model (\textsc{R-pred}) predicts the most likely temporal unit/range for the event, with a classification output layer; and the other model (\textsc{E-pred}) predicts the \textit{exact} duration value, with a regression output layer.
Our best model (\textsc{E-pred}) achieves state-of-the-art performance on the TimeBank dataset and the McTACO duration prediction task. 
In addition, in the unsupervised setting, our model (\textsc{E-pred}) trained with only collected web data outperforms the supervised BERT baseline by 10.24 F1 score and 9.68 Exact Match score on McTACO duration prediction task. 
We also provide detailed comparisons and analysis between the regression objective (\textsc{E-pred}) and the classification objective (\textsc{R-pred}).

\section{Our Framework} 

\subsection{Duration-relevant Sentences Collection and Automatic Labeling}\label{Collecting data}

We use multiple pattern-based extraction rules to collect duration-relevant sentences. To avoid the potential data sparsity problem, we extract them from a relatively large corpus. 
In particular, we use articles in DeepMind Q\&A dataset~\cite{nips15_hermann} which contains approximately 287k documents extracted from CNN and Daily Mail news articles.
To avoid introducing potential bias from a single pattern, we design multiple patterns for extraction. Specifically, if a sentence contains words or its variants as ``for'', ``last'', ``spend'', ``take'', ``over'', ``duration'', ``period'', and within certain number of words there exists a numerical value and a temporal unit (including second, minute, hour, day, week, month, year, decade) 
, then we consider the sentence as containing duration information and keep the sentence. Further, we design rules to filter sentences with certain patterns to avoid common misjudgements of the patterns to reach higher precision in retrieving sentences with duration information. More details are illustrated in Appendix~\ref{appendix_data_collection}.

\begin{figure}[t]
\centering
\resizebox{\columnwidth}{!}{
\includegraphics{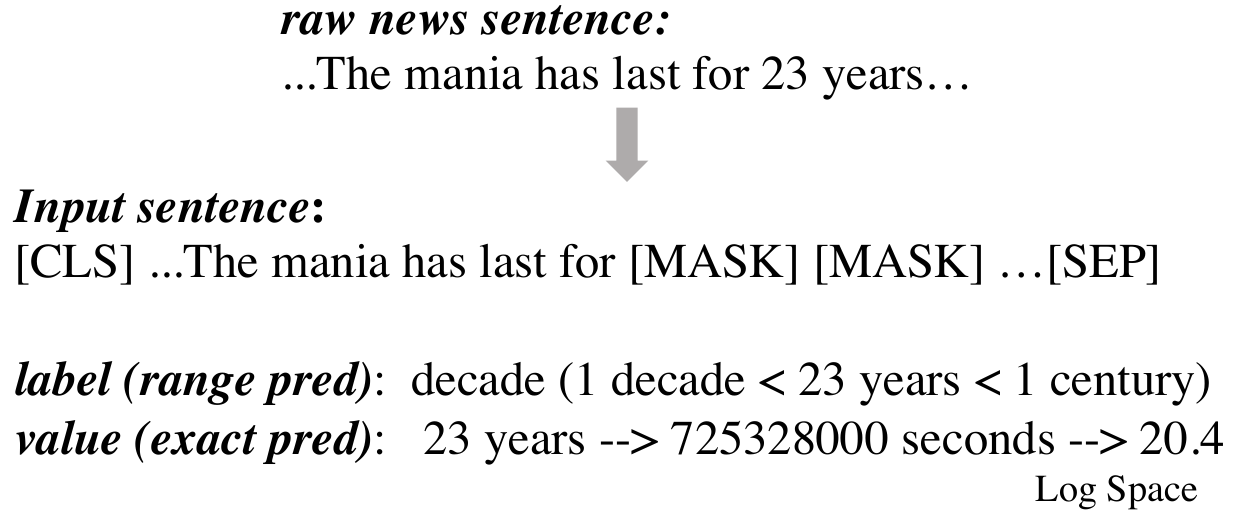}
}
\caption{An Example of Automatic Labeling}
\label{autmatic_labeling}
\end{figure}

We apply rules to create the labels (Figure~\ref{autmatic_labeling}),
specifically, given a candidate sentence, we extract the duration expression (23 years) which consists of a number and unit, then we normalize it to ``second'' space. We use the logarithm of the normalized value as label for \textsc{E-pred}; and use the closest temporal unit  as label for \textsc{R-pred} model. Then for the sentence itself, we replace its duration expression with [MASK]s.

\subsection{Models for Duration Prediction} \label{method and pretraining}
\begin{figure}[t]
\centering
\resizebox{\columnwidth}{!}{
\includegraphics[scale=0.55]{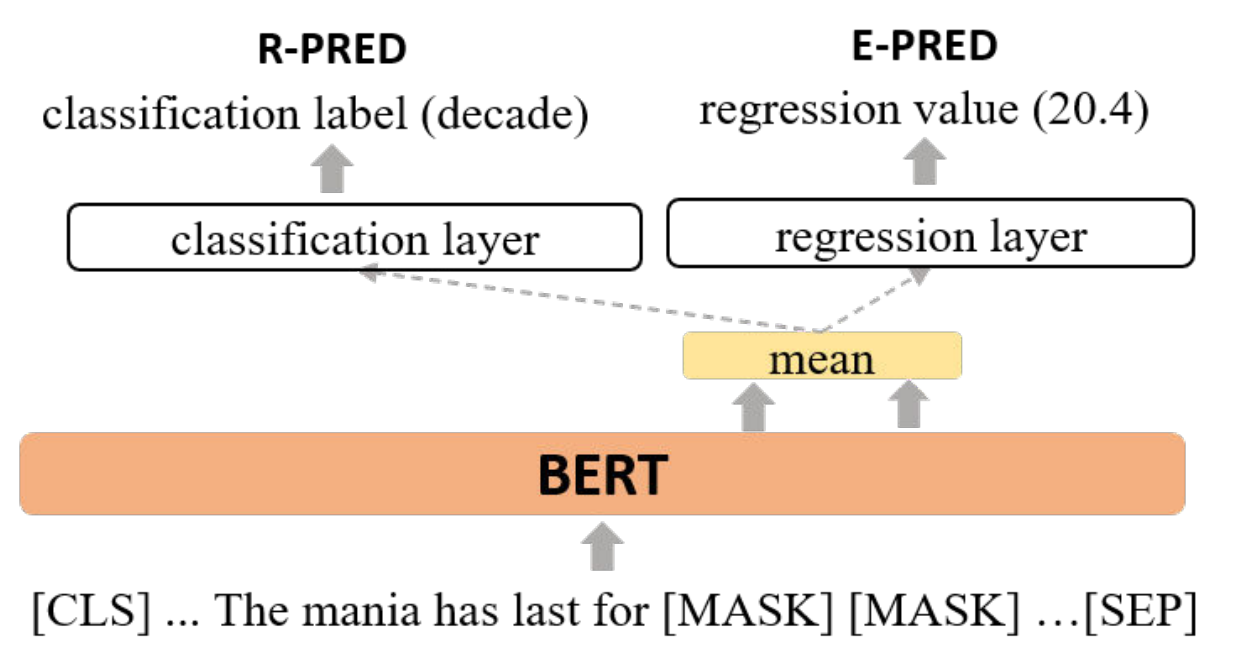}
}
\caption{Models: \textsc{R-pred} and \textsc{E-pred}}
\label{fig:Model 1}
\end{figure}

The structure of \textsc{E-pred} and \textsc{R-pred} is shown in Figure~\ref{fig:Model 1}.
We first pass the input sentence through BERT~\cite{devlin-etal-2019-bert} to obtain contextualized embedding for the masked tokens, $\mathbf{x}_0$, $\mathbf{x}_1$, ..., $\mathbf{x}_k$. 
Then we add a linear layer on top of the BERT representations for prediction. We propose two variations -- 
\textsc{E-pred} (with a regression layer) predicts the exact duration value $v$; 

\begin{equation}
\label{equation_duration_value}
\small
\begin{gathered}
\nonumber
    \mathbf{v} = \mathbf{W}_e\sum_{i=0}^{k}{\mathbf{x}_i} \\
\end{gathered}
\end{equation}

\textsc{R-pred} (with a cross-entropy layer) predicts the range $r$.
\begin{equation}
\label{equation_duration_value}
\small
\begin{gathered}
\nonumber
    \mathbf{r} = \text{softmax}(\mathbf{W}_r\sum_{i=0}^{k}{\mathbf{x}_i})
\end{gathered}
\end{equation}


\section{Experiments and Analysis} \label{experiments}

\begin{table*}[t]
\resizebox{\textwidth}{!}{
\begin{tabular}{lcccccccc}
\toprule
\multicolumn{1}{l|}{}                        & \multicolumn{3}{c|}{Coarsed-Grained (Test)}                                    & \multicolumn{3}{c|}{Coarsed-Grained (TestWSJ)}                                 & \multicolumn{2}{c}{Fine-Grained}                 \\
\multicolumn{1}{l|}{\multirow{-2}{*}{Model}} & \textless{}day F1 & \textgreater{}day F1 & \multicolumn{1}{l|}{Acc.}       & \textless{}day F1 & \textgreater{}day F1 & \multicolumn{1}{l|}{Acc.}       & Acc. (Test)               & Acc. (TestWSJ) \\ 
\midrule
\multicolumn{9}{c}{Supervised Setting}  \\
\midrule
\multicolumn{1}{l|}{Majority class}          & -                 & 76.90                & \multicolumn{1}{c|}{62.47}          & -                 & 76.99                & \multicolumn{1}{c|}{62.58}          & 59.28                        & 52.38             \\
\multicolumn{1}{l|}{Maximum Entropy \cite{pan-etal-2011-annotating}$^\dagger$}                     & -                 & -                    & \multicolumn{1}{c|}{73.30}          & -                 & -                    & \multicolumn{1}{c|}{73.50}          & 62.20                        & 61.90             \\
\multicolumn{1}{l|}{Maximum Entropy++ \cite{gusev2011using}$^\dagger$}                   & -                 & -                    & \multicolumn{1}{c|}{73.00}          & -                 & -                    & \multicolumn{1}{c|}{74.80}          & 62.40                        & 66.00             \\
\multicolumn{1}{l|}{LSTM ensemble \cite{vempala-etal-2018-determining}}           & 64.29             & 82.69                & \multicolumn{1}{c|}{76.69}          & 73.20              & 87.78                & \multicolumn{1}{c|}{83.21}          & -                            & -                 \\
\multicolumn{1}{l|}{TACOLM \cite{zhou-etal-2020-temporal}}                  & 80.58             & 88.88                & \multicolumn{1}{c|}{85.86}          & \textbf{76.01}    & \textbf{88.14}       & \multicolumn{1}{c|}{\textbf{84.12}} & -                            & -                 \\ 
\midrule
\multicolumn{1}{l|}{\textsc{R-PRED}}                  & \textbf{82.08}    & 87.72                & \multicolumn{1}{c|}{85.43}          & 70.15             & 81.12                & \multicolumn{1}{c|}{76.87}          & 82.09 & 76.19             \\
\multicolumn{1}{l|}{\quad w/o pre-training}        & 80.94             & 86.19                & \multicolumn{1}{c|}{84.01}          & 73.46             & 79.93                & \multicolumn{1}{c|}{77.32}          & 80.38                        & \textbf{78.46}    \\
\multicolumn{1}{l|}{\textsc{E-PRED}}                  & 80.63             & \textbf{89.46}       & \multicolumn{1}{c|}{\textbf{86.35}} & 70.67             & 85.39                & \multicolumn{1}{c|}{80.50}          & \textbf{82.52}               & \textbf{78.46}    \\
\multicolumn{1}{l|}{\quad w/o pre-training}        & 78.73             & 88.16                & \multicolumn{1}{c|}{84.79}          & 73.50             & 86.21                & \multicolumn{1}{c|}{81.86}          & 80.34                        & 77.02             \\ 
\midrule
\multicolumn{9}{c}{Unsupervised Setting}                                                                                                                                                                                                                          \\ 
\midrule
\multicolumn{1}{l|}{Majority}                & -                 & 76.90                & \multicolumn{1}{c|}{62.47}          & -                 & 76.99                & \multicolumn{1}{c|}{62.58}          & 59.28                        & 52.38             \\
\multicolumn{1}{l|}{Web count, yesterday \cite{gusev2011using}$^\dagger$}    & -                 & -                    & \multicolumn{1}{c|}{70.70}          & -                 & -                    & \multicolumn{1}{c|}{\textbf{74.80}} & -                            & -                 \\
\multicolumn{1}{l|}{Web count, bucket \cite{gusev2011using}$^\dagger$}       & -                 & -                    & \multicolumn{1}{c|}{72.40}          & -                 & -                    & \multicolumn{1}{c|}{73.50}          & 66.50                       & \textbf{68.70}   \\ 
\midrule
\multicolumn{1}{l|}{\textsc{R-PRED}}                  & 63.19             & 80.39                & \multicolumn{1}{c|}{74.41}          & 5.19              & 66.36                & \multicolumn{1}{c|}{50.34}          & 69.72                        & 43.54             \\
\multicolumn{1}{l|}{\textsc{E-PRED}}                  & 60.14             & 82.52                & \multicolumn{1}{c|}{\textbf{75.69}} & 2.86              & 69.64                & \multicolumn{1}{c|}{53.74}          & \textbf{71.00}               & 41.50             \\ 
\bottomrule
\end{tabular}
}
\caption{Performance on TimeBank. Results marked with $\dagger$ are reported in \newcite{gusev2011using}.}
\label{timebank_experiments} 
\end{table*}

\subsection{Datasets and Evaluation Metrics}
We evaluate our models on two duration-prediction benchmarks -- 
TimeBank~\cite{pan-etal-2011-annotating} and 
McTACO-duration~\cite{zhou-etal-2019-going}. 
%
\textbf{TimeBank}\footnote{We use \newcite{gusev2011using}'s split and obtain 1663/469/147 events in Train/Test/TestWSJ set respectively.}
annotates 48 non-Wall-Street-Journal articles (non-WSJ) and 10 WSJ articles. 
Specifically, it annotates duration for an event trigger 
(e.g., ``watched'') 
in the sentence 
(e.g., I \underline{watched} a movie yesterday). 
Non-WSJ articles are splitted to generate train set and test set, and WSJ articles are used to generate testWSJ set, serving as an additional evaluation set.
The Coarse-Grained task requires predicting whether the event takes less than a day or longer than a day; the Fine-Grained task requires predicting the most likely temporal unit (e.g., second, minute, hour, etc.).
To transform the sentences into the input format of our models. We insert duration pattern (``, lasting [MASK] [MASK], '') after event word and use the new sentence as the input sequence.
For example, one sentence in TimeBank is ``Philip Morris Cos, \underline{adopted} a defense measure ...''. Our method will convert it to ``Philip Morris Cos, adopted, lasting [MASK] [MASK], a defense measure ...''. Our strategy of directly adding duration pattern is possible to help pre-trained model to utilize learned intrinsic textual representation for duration prediction~\cite{tamborrino-etal-2020-pre}.

McTACO is a multi-choice question answering dataset.  \textbf{McTACO-duration}\footnote{In practice we collect context-question-answer triples that questions are about event duration and answers can be transformed to a duration value. We get 1060/2827 triples for dev/test set respectively (out of 1112/3032).}
is a subset of McTACO whose questions are about event duration. 
Each data item includes a context sentence, a question, an answer (a duration expression) and a label indicating whether the answer is correct or not. 
To obtain the input sequence for our model,
we convert the question to a statement using rule based method, and insert the same ``, lasting [MASK] [MASK].'' to the \textit{end} of the statement sentence.
For example, one question in McTACO-duration is “How long would they run through the fields?”, our method will convert it to “they run through the fields, lasting [MASK] [MASK].”
We then join the context sentence and newly obtained statement sentence as the input sequence. 

We report F1 and accuracy for TimeBank Coarse-Grained task and accuracy for TimeBank Fine-Grained task. We report F1 and Exact Match (EM) for McTACO-duration.


\subsection{Additional Dataset Details}

In TimeBank Coarse-grained task, given an input event sentence, if prediction of \textsc{E-pred} is smaller than 86400 seconds or prediction of \textsc{R-pred} is ``second'' or ``minute'' or ``hour'', prediction will be ``< day''; Otherwise prediction will be ``> day''.
All models in TimeBank Fine-Grained task uses approximate agreement \cite{pan-etal-2011-annotating} during evaluation. In approximate agreement, temporal units are considered to match if they are the same temporal unit or adjacent ones. For example, ``second'' and ``minute'' match, but ``minute'' and ``day'' do not. It is proposed because human agreement on exact temporal unit is low (44.4\%).

For McTACO-duration task, \textsc{E-pred} uses $range$ as a hyper-parameter to define whether the answer is correct or not. Specifically, if the prediction of \textsc{E-pred} is $d$, then only answers in $d \pm range$ in logarithmic second space are predicted as correct. We tune $range$ in development set. Here the $range$ we use is 3.0. \textsc{R-pred} uses approximate agreement to predict correctness.

\subsection{Baselines}
We compare to strong models in the literature.
For TimeBank, 
\textbf{Majority Class} always select ``month'' as prediction (``week'', ``month'' and ``year'' are all considered as match because of approximate agreement).
In the supervised setting,
\textbf{Maximum Entropy}~\cite{pan-etal-2011-annotating} and \textbf{Maximum Entropy++}~\cite{gusev2011using} are two models which utilize hand-designed time-related features. Difference is that Maximum Entropy++ uses more features than Maximum Entropy.
\textbf{LSTM ensemble}~\cite{vempala-etal-2018-determining} is an ensemble LSTM~\cite{hochreiter1997long} model which utilize word embeddings.
\textbf{TACOLM}~\cite{zhou-etal-2020-temporal} is a concurrent work to our methods that also utilize unlabeled data. It uses a transformer-based structure and is also pre-trained on automatically labeled temporal sentences.
Different from our model, TACOLM focuses on classification model and providing better representation instead of directly generating predicted duration. 
Here TACOLM forms Coarse-Grained task as a sequence classification task and uses the embedding of the first token of transformer output to predict from ``< day'' or ``> day''.

For McTACO-duration, 
\textbf{BERT\_QA}~\cite{zhou-etal-2019-going} is the BERT sentence pair (question and answer) classification model trained with McTACO-duration;
\textbf{BERT\_QA full} is the same model trained with all of McTACO examples.
\textbf{TACOLM} here shares the same structure with BERT\_QA but uses transformer weights pre-trained on collected data.
To be fair, train data for TACOLM is McTACO-duration, the same as \textsc{R-pred} and \textsc{E-pred}. 
For the unsupervised setting,
for TimeBank, we compare to \textbf{Web count-yesterday} and \textbf{Web count-bucket}~\cite{gusev2011using}. They are rule-based approaches which rely on search engine.
%

\subsection{Results}

Table~\ref{timebank_experiments} presents results for TimeBank. 
In the supervised setting \textsc{E-pred} achieves the best performance in Coarse-Grained task (``Test set'') and Fine-Grained task, while it receives a lower performance than TACOLM in Coarse-Grained task (``TestWSJ''). In addition, \textsc{E-pred} achieves best performance in Test set in unsupervised setting while it receives lower performance in TestWSJ set. However, Test set has a similar distribution with train set, while TestWSJ's is different (from a different domain). Therefore, performance on Test set should be a more important indicator for comparison. 

We attribute the possible limitation of our models in TimeBank (especially TestWSJ set) experiments to reporting bias, relatively limited number of automatically collected data and mismatch of our duration pattern and TimeBank style annotation. More details are explained in Section \ref{analysis_pretraining}. TACOLM's better performance in Coarse-Grained task in TestWSJ set might caused by its more compatible input format with TimeBank (it marks each event word that has a duration annotation in collected data) and its larger number of collected data from more sources.

%

%

Table~\ref{mctaco_experiments} presents result on McTACO-duration. In supervised setting, \textsc{E-pred} achieves the best performance. This table indicates that pre-training for incorporating external textual knowledge is helpful for both \textsc{R-pred} and \textsc{E-pred}. Plus, \textsc{E-pred} which is trained with \textit{only} web collected data still outperforms BERT\_QA by a large margin.

We observe that \textsc{E-pred} and \textsc{R-pred} does not receive much performance gain from task-specific training. We attribute it to the noise introduced during transforming the QA data to fit in our models' input-output format. 
Specifically, we use the average of all correct answers as duration value label. This process is not guaranteed to get the expected duration value for each input event sentence.


\begin{table}[t] 
\centering
\small
\begin{tabular}{lcc}
\toprule
Model                                                                       & \multicolumn{1}{c}{F1} & \multicolumn{1}{c}{EM} \\ \midrule
\multicolumn{3}{c}{Supervised setting}          \\ \midrule
BERT\_QA        & 51.95                  & 30.32                  \\
BERT\_QA full  & 56.98                  & 32.26                  \\
TACOLM \cite{zhou-etal-2020-temporal}  & 57.60  & 33.50 \\
\midrule
\textsc{R-pred}                                    & 55.36                  & 25.48                  \\
\quad w/o pre-training         & 50.05                  & 22.58                  \\
\textsc{E-pred}                                        & \textbf{63.63$^*$}         & \textbf{39.68$^*$}         \\
\quad w/o pre-training                                                               & 45.31                  & 25.48                  \\ \midrule
\multicolumn{3}{c}{Unsupervised Setting}                                                                                      \\ \midrule
\textsc{R-pred}                                             & 54.14                  & 25.16                  \\
\textsc{E-pred}                                                 & \textbf{62.19}         & \textbf{40.00}         \\ \bottomrule
\end{tabular}

\caption{Performance on McTACO-duration. * indicates that the difference compared to BERT\_QA is statistically significant ($p<0.01$) using Bootstrap method \cite{berg2012empirical}}
\label{mctaco_experiments} 
\end{table}

\subsection{Analysis}

\paragraph{\textsc{E-pred} or \textsc{R-pred}?}
We provide insights on why BERT with regression loss generally outperforms BERT with a classification loss.

Firstly, we observe empirically that \textsc{E-pred} generally outperforms \textsc{R-pred} in TimeBank experiments. We attribute that \textsc{E-pred} can catch more nuance information than \textsc{R-pred}. For example, if the duration mentioned in the text is 40 min, then the generated label for \textsc{R-pred} is ``minute''. While for \textsc{E-pred}, the generated label is 40 minutes (1 min v.s. 40 min).

Secondly, \textsc{E-pred} is more flexible and have a tunable range to predict the correctness (one of main reasons that \textsc{E-pred} outperforms \textsc{R-pred} largely in Table~\ref{mctaco_experiments}), while \textsc{R-pred} can only use single bucket prediction or approximate agreement. 


\begin{figure}[t] 
\centering
\resizebox{\columnwidth}{!}{
\includegraphics{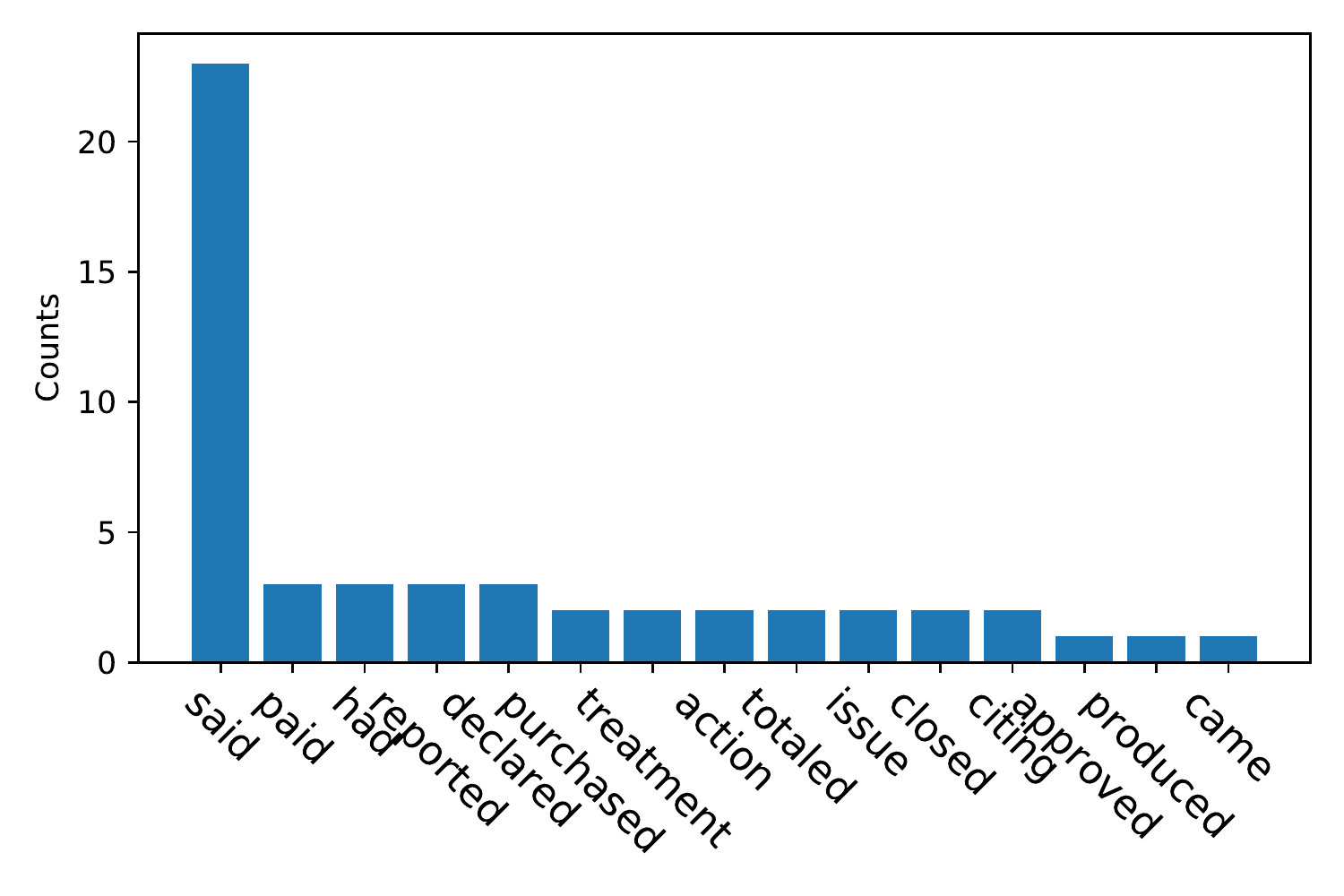}
}
\caption{Times of event words that are predicted incorrectly by \textsc{E-pred} in TimeBank TestWSJ set in unsupervised setting (only showing the 15 most frequent event words).}
\vspace{-0.15cm}
\label{TBWSJ_incorrect_prediction_statistics}
\end{figure}

\paragraph{Effect of Time-aware Pre-training}
\label{analysis_pretraining}
We observe that time-aware pre-training can lead to 5\textasciitilde{18} F1 score improvement in McTACO-duration; while in TimeBank Coarse-Grained task, it can only lead to 1\%\textasciitilde{3\%} accuracy improvement in Test set, and causes around 1\% accuracy drop in TestWSJ set.

We attribute the relatively limited effect of time-aware pre-training in TimeBank to reporting bias \citep{gordon2013reporting} and data difference between McTACO-duration and TimeBank. 
Specifically, annotated events in McTACO-duration are mainly description of concrete events,
while annotated events in TimeBank are mainly abstract single words in the sentence.
We consider that events in McTACO are more similar to events in our automatically collected data, while events in TimeBank are far less similar. 
Specifically, Figure~\ref{TBWSJ_incorrect_prediction_statistics} shows the most frequent single words annotated in TestWSJ that are predicted wrongly by \textsc{E-pred} in unsupervised setting. 
We observe that event words in Figure~\ref{TBWSJ_incorrect_prediction_statistics} are mainly abstract and not durative, and people usually do not describe the duration of them in text (reporting bias). However, a larger collection of automatically collected data from different sources might alleviate this problem. More details on
error analysis
in TimeBank experiments can be found in Appendix~\ref{Appen_TB_analysis}.

Another reason could be the mismatch of our designed duration pattern and TimeBank annotation style. Directly adding duration pattern after the annotated word might not comply with the sentences seen in pre-training data and might cause ambiguous reference of event.


\paragraph{Influence of Data Collection and Search Patterns}
We investigate how pre-training data collection affects the performance of our models. Table~\ref{table_data_collection_methods} shows performance of \textsc{E-pred} in unsupervised setting pretrained w/ data collected with different methods. 
Specifically, we collect duration sentences from News or Wikipedia articles; sentences are collected by only the ``for'' pattern or ``for$|$take$|$spend$|$last$|$lasting$|$duration$|$period'' patterns (7 patterns). 
We find that \textsc{E-pred} pre-trained with the three data collecting methods all achieves state-of-the-art performance in TimeBank Test (unsupervised setting) and get higher F1 score than BERT\_QA supervised baseline.
We find that pre-training with collected sentences can robustly increase our model's understanding of duration, and using more patterns for data collection is beneficial.

\begin{table}[t]
\small
\centering
\resizebox{\columnwidth}{!}{
\begin{tabular}{l|cc|cc}
\toprule
                    & \multicolumn{2}{c|}{TimeBank}   & \multicolumn{2}{c}{McTACO-duration} \\ \midrule
                    & Test           & TestWSJ        & F1               & EM               \\
Wiki (7 patterns)   & 70.15          & \textbf{46.26} & 57.34            & 36.77            \\
News (only ``for'') & 67.80          & 43.54          & 58.89            & 36.77            \\
News (7 patterns)   & \textbf{71.00} & 41.50          & \textbf{62.19}   & \textbf{40.00}   \\ \bottomrule
\end{tabular}
}
\caption{Effect of Data Collection and Search Patterns.}
\vspace{-0.1cm}
\label{table_data_collection_methods}
\end{table}

\section{Additional Related Work}
For \textbf{supervised} duration prediction, 
\newcite{pan-etal-2011-annotating} annotates duration length of a subset of events in TimeBank~\cite{pustejovsky2003timebank}. 
New features and learning based models are proposed for TimeBank ~\cite{pan-etal-2011-annotating, gusev2011using, samardzic2016aspectbased, vempala-etal-2018-determining}. In particular, aspectual~\cite{vendler1957verbs, smith2013parameter} features have been proved to be useful.
%
Concurrent to our work, \newcite{zhou-etal-2020-temporal} also utilize unlabeled data.
Different from our work, they focus on temporal commonsense \textit{acquisition} in a more general setting (for frequency, typical time, duration, etc.) and the models predict the discrete temporal unit, while we propose two models (classification and regression-based). In addition, they focus on providing better representation instead of directly generating duration prediction. 
For the \textbf{unsupervised} setting, \newcite{williams-katz-2012-extracting, elazar-etal-2019-large} use rule-based method on web data and generate collections of mapping from verb/event pattern to numeric duration value. \newcite{kozareva2011learning, gusev2011using} develop queries for search engines and utilize the returned snippets / hitting times to make prediction.

\section{Conclusion}
We propose a framework for leveraging free-form textual knowledge into neural models for duration prediction.
Our best model (\textsc{E-pred}) achieves state-of-the-art performance in various tasks.
In addition, our model trained only with externally-obtained weakly supervised news data outperforms supervised BERT\_QA baseline 
by a large margin. 
We also find that model trained with exact duration value seems to better capture duration nuance of event, and 
has more tunable range that is more flexible to make prediction for quantitative attributes such as duration.

\section*{Acknowledgments}
We thank the anonymous reviewers for suggestions and Ben Zhou for running experiment of TACOLM on McTACO-duration dataset.

\bibliography{anthology, emnlp2020}
\bibliographystyle{acl_natbib}

\cleardoublepage
\newpage
\appendix

\section{Appendices}
\label{sec:appendix}
\subsection{Hyper-Parameters} \label{appendix_hyperparameters}
For pre-training BERT model with collected cheap supervised data, we use the same hyper parameters for time aware \textsc{R-pred} and \textsc{E-pred}: 

\begin{itemize}
\item learning rate: 5e-5 
\item train batch size: 16
\item optimizer: BertAdam (optimizer warmup proportion: 0.1)
\item loss: mean square error loss (for \textsc{E-pred}); cross entropy loss (for \textsc{R-pred}) 
\end{itemize} 

For fine-tuning \textsc{R-pred} or \textsc{E-pred} with McTACO-duration or TimeBank data or fine-tuning BERT with McTACO-duration or TimeBank data, the hyper-parameter we use is: 

\begin{itemize}
\item learning rate: 2e-5
\item train batch size: 32
\item optimizer: BertAdam (optimizer warmup proportion: 0.1)
\item loss: mean square error loss (for \textsc{E-pred}); cross entropy loss (for \textsc{R-pred}) 
\end{itemize} 

\subsection{Duration Data Collecting Method} \label{appendix_data_collection}
We firstly use regular expression pattern to retrieve sentences that match with the pattern, then we use filter patter to filter out sentences that match with filter out pattern. 

Regular expression pattern:
``(?:duration$|$period$|$for$|$last$|$lasting$|$spend
$|$spent$|$over$|$take$|$took$|$taken)[$\land$,.!?;]*$\backslash$d+ (?:second$|$minute$|$hour$|$day$|$week$|$month$|$year$|$decade)''

Filter pattern:
\begin{itemize}
\item if the matched sub-sentence contains ``at'' or ``age'' or ``every'' or ``next'' or ``more than'' or ``per''
\item if the matched sub-sentence match with ``(?:first$|$second$|$third$|$fourth$|$fifth$|$sixth$|$seventh
$|$eighth$|$ninth) time''
\item if the matched sentence matches with ``$|$d+ secondary''
\item if the matched sentence matches with ``(?:second$|$minute$|$hour$|$day$|$week$|$month$|$year
$|$decade)[s]? old''
\end{itemize} 





\subsection{Additional Details on Processing TimeBank and McTACO Data}
Each annotated event trigger word in TimeBank are labeled with two duration values, max duration and min duration. We use the arithmetic mean of the two values to generate labels.

For TimeBank Fine-grained task, we use 7 temporal units as all possible labels (same setting with previous work \cite{gusev2011using} \cite{pan-etal-2011-annotating}), including ``second'', ``minute'',  ``hour'',  ``day'',  ``week'',  ``month'', ``year''. For \textsc{R-pred} in McTACO task, we use 8 temporal units instead (adding ``decade'')

\subsection{Details on Correctly and Incorrectly Predicted Event Words in TimeBank Experiment}
\label{Appen_TB_analysis}

\begin{figure}[t]
\centering
\resizebox{\columnwidth}{!}{
\includegraphics{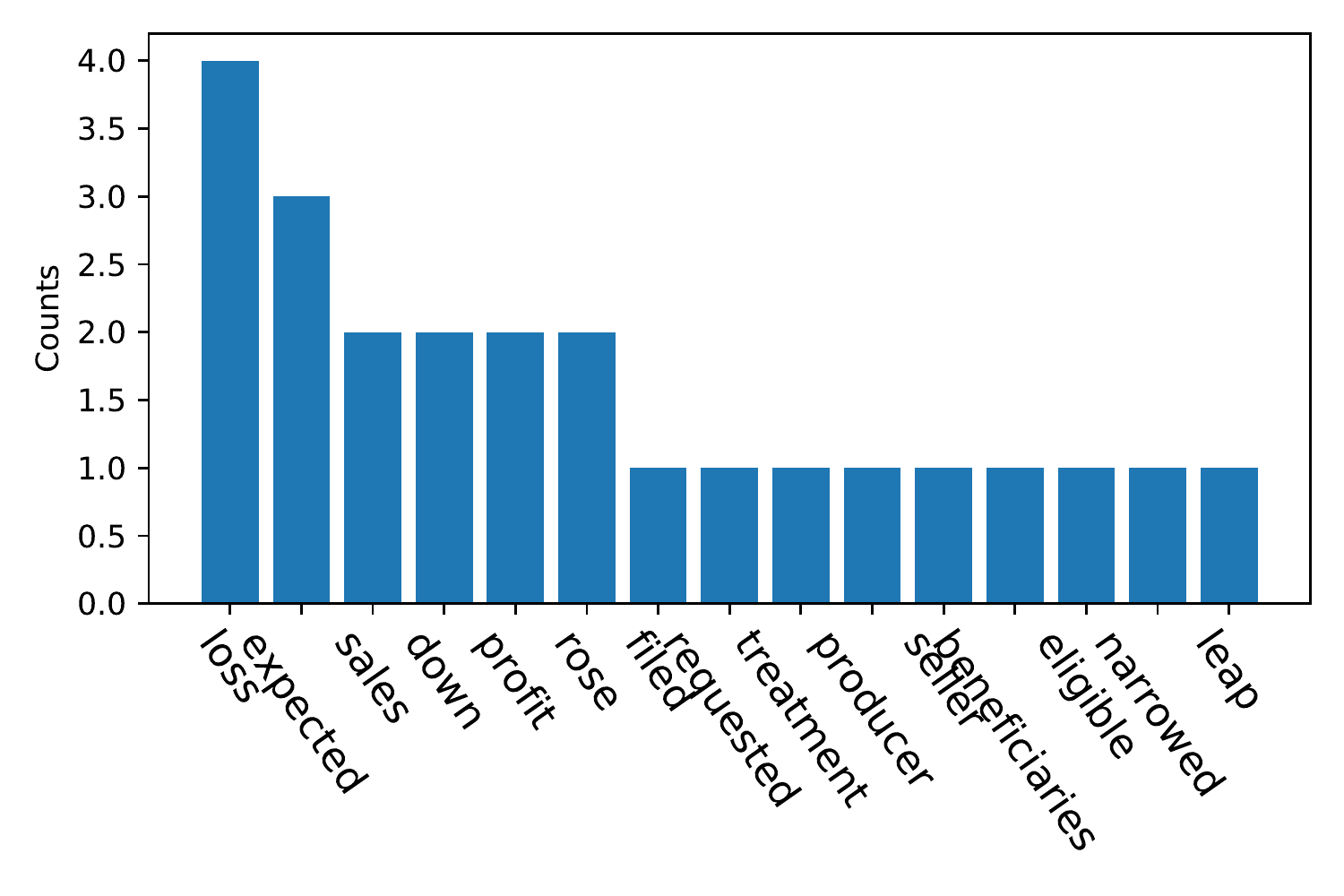}
}
\caption{Times of event words that are predicted correctly in TimeBank TestWSJ set in unsupervised setting (only shows most frequent 15 event words)}
\vspace{-0.15cm}
\label{TBWSJ_correct_prediction_statistics}
\end{figure}

\begin{figure}[t]
\centering
\resizebox{\columnwidth}{!}{
\includegraphics{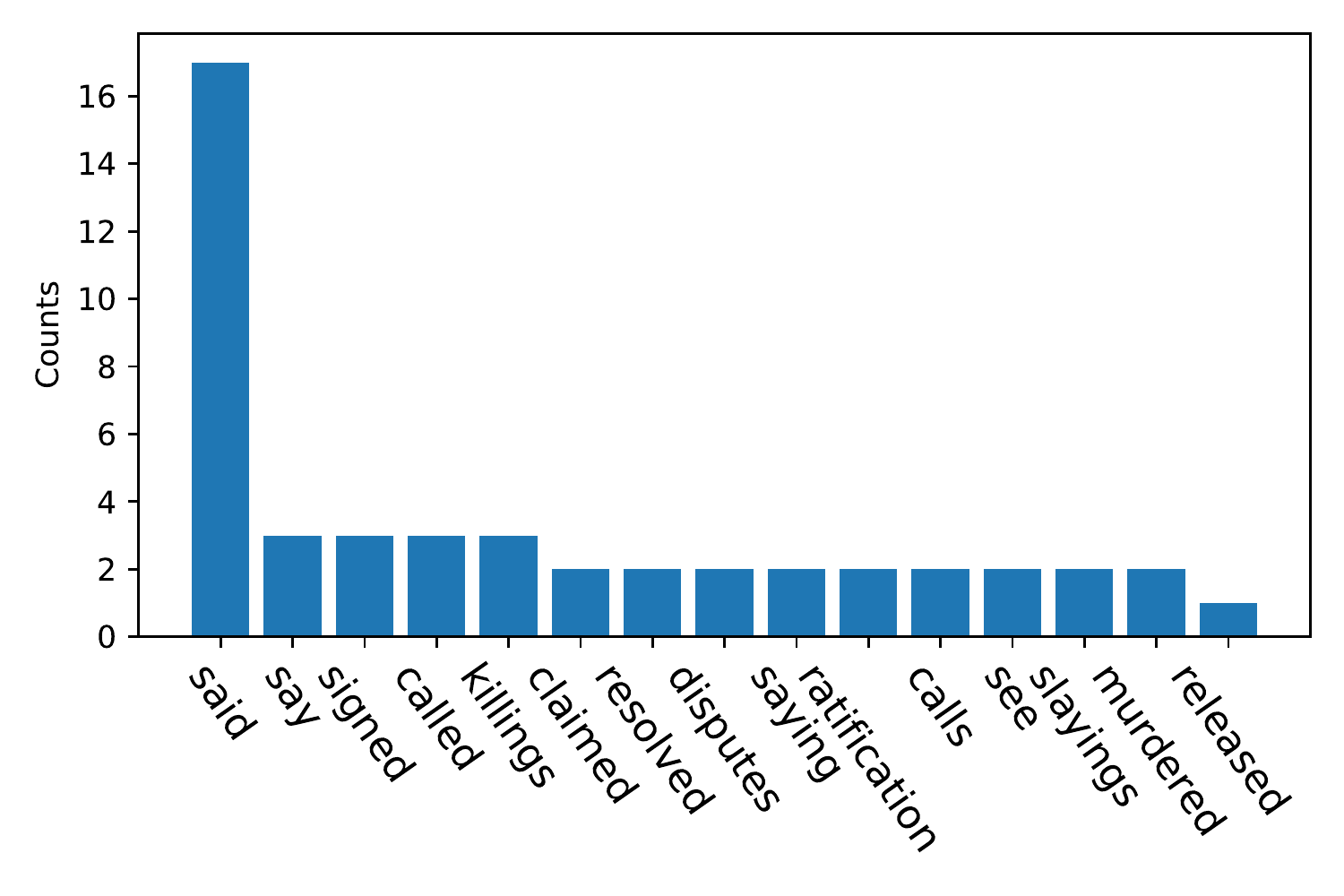}
}
\caption{Times of event words that are predicted incorrectly in TimeBank Test set in unsupervised setting (only shows most frequent 15 event words)}
\vspace{-0.15cm}
\label{TBtest_incorrect_prediction_statistics}
\end{figure}

\begin{figure}[t]
\centering
\resizebox{\columnwidth}{!}{
\includegraphics{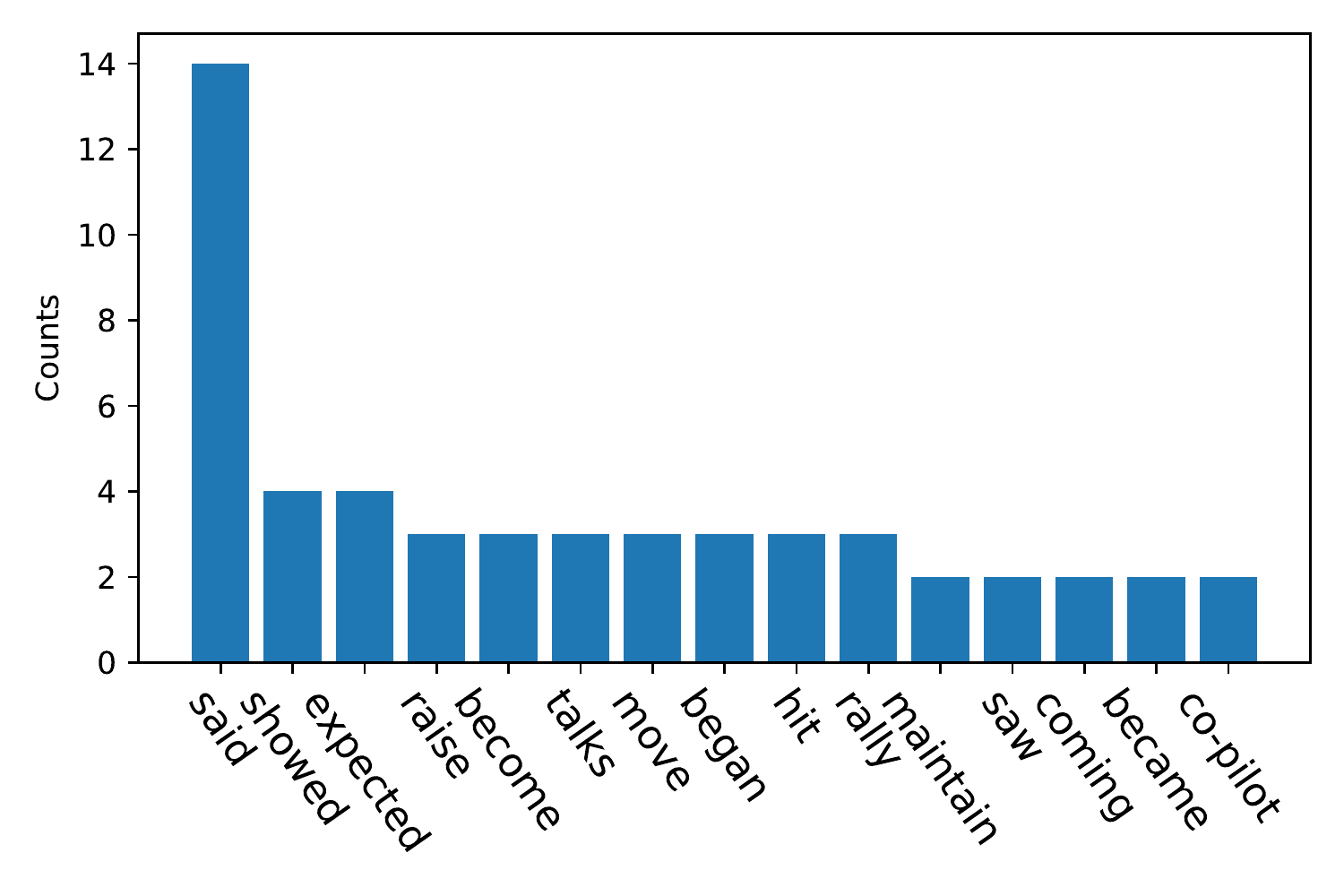}
}
\caption{Times of event words that are predicted correctly in TimeBank Test set in unsupervised setting (only shows most frequent 15 event words)}
\vspace{-0.15cm}
\label{TBtest_correct_prediction_statistics}
\end{figure}

As shown in Figure~\ref{TBWSJ_correct_prediction_statistics},
Figure~\ref{TBtest_incorrect_prediction_statistics} and 
Figure~\ref{TBtest_correct_prediction_statistics}, we observe that correctly predicted words are generally more concrete and more possible to be described duration in text, which supports our analysis on reporting bias.

\end{document}